# GVdoc: Graph-based Visual Document Classification

**Fnu Mohbat**[1]*, **Mohammed J. Zaki**[1], **Catherine Finegan-Dollak**[2†], **Ashish Verma**[3†]
[1]Rensselaer Polytechnic Institute, [2]University of Richmond, [3]Amazon
mohbaf@rpi.edu, zaki@cs.rpi.edu,
cfinegan@richmond.edu, draverma@amazon.com

## Abstract

The robustness of a model for real-world deployment is decided by how well it performs on unseen data and distinguishes between in-domain and out-of-domain samples. Visual document classifiers have shown impressive performance on in-distribution test sets. However, they tend to have a hard time correctly classifying and differentiating out-of-distribution examples. Image-based classifiers lack the text component, whereas multi-modality transformer-based models face the token serialization problem in visual documents due to their diverse layouts. They also require a lot of computing power during inference, making them impractical for many real-world applications. We propose, GVdoc, a graph-based document classification model that addresses both of these challenges. Our approach generates a document graph based on its layout, and then trains a graph neural network to learn node and graph embeddings. Through experiments, we show that our model, even with fewer parameters, outperforms state-of-the-art models on out-of-distribution data while retaining comparable performance on the in-distribution test set.

## 1 Introduction

Documents digitization and their intelligent processing in various industries such as finance, insurance, and medicines has resulted in the rapid development of structured document understanding methods, a.k.a. document AI. Document classification is one of the essential tasks in document AI for labeling documents. A number of deep convolutional neural network (CNN) and Transformer-based models have achieved superior performance on many document-AI tasks (Xu et al., 2021; Lee et al., 2021, 2022). However, they tend to employ bigger models with hundreds of millions of parameters, subsequently increasing computational demand that can be a challenge in real-world applications. Yet many of them fail to perform well on out-of-distribution (OOD) data (Larson et al., 2021, 2022). This is because, in many cases, training and testing examples are from a fixed distribution − such as a particular language, time frame, and industry. However, the layout of the documents evolves over time, and the model should perform well on such out-of-distribution data. Further, the model is expected to be able to differentiate between known and unknown categories of documents, thus minimizing false-positive predictions during testing.

Initial work on document classification employed off-the-shelf image classifiers (Jain and Wigington, 2019; Bakkali et al., 2020) and models pre-trained on ImageNet (Deng et al., 2009) or similar datasets. These methods struggle to label documents having similar layouts but different text contexts. Later, focus shifted towards language models (Li et al., 2021a; Lee et al., 2022) and multi-modality models (Bakkali et al., 2020; Xu et al., 2021; Lee et al., 2021; Wang et al., 2022a). These models also incorporated layout information obtained from optical character recognition (OCR). Therefore, the performance of these methods, particularly transformer-like models, degrades due to the imperfection of the OCR engine, such as errors in parsed text or the order of tokens sequence. Almost all of these methods tried to improve the performance on the in-distribution test set, neglecting the generalization for real-world applications. To confirm, recently (Larson et al., 2022) collected an OOD version of RVLCDIP dataset (Harley et al., 2015) and evaluated several image and multi-modal classifiers. However, none of them performed well on the OOD dataset.

Our method, called GVdoc (for **G**raph-based **V**isual **DO**cument **C**lassification), studies docu-ment classification as a graph classification problem, where we take text words as nodes and the relationship between words as edges in a graph. We generate a document-level graph using that layout information from OCR (see Figure 1) and learn the embedding using graph neural networks (GNNs). GVdoc is more robust to changes in the test set; hence it shows improved performance on out-of-distribution data. We make the following contributions:

- We introduce graph-based document modeling that leverages both (potentially noisy) reading order and spatial layout in graph construction, and learns embeddings using GNNs.

- We empirically show that compared with other systems, our model is better able to generalize to test data drawn from a different distribution than the training data.

---

*The work was partially done during an externship at IBM T. J. Watson Research Center.
† was at IBM T. J. Watson Research Center.

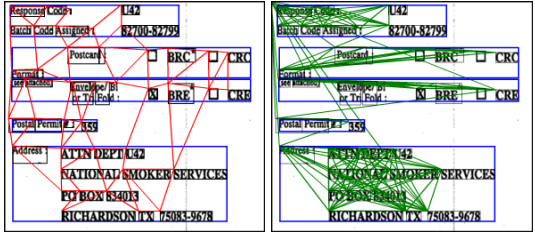

Figure 1: Sample document graph where the bounding boxes of words are shown by black boxes, and paragraphs by blue boxes. Left side figure shows $\beta$ skeleton edges with red lines and right side shows OCR-based paragraph-level edges with green color. The edges from left top corner connect the super node to some representative nodes. The final graph is combination of both of these graphs (see Figure 11 in Appendix).

## 2 Related Work

**Visual Document Classification** CNNs have achieved excellent performance on natural scene images, so they became the first obvious choice for visual document classification (Das et al., 2018; Jain and Wigington, 2019; Bakkali et al., 2020). However, documents have overlapping intra-class visual and structural characteristics (Bakkali et al., 2020), which makes visual features less discriminative for classification. The semantics of text in the document and the layout are essential to understand the visual documents.

A second line of work studies document classification as a sequence classification problem (Lee et al., 2022; Li et al., 2021a; Wang et al., 2022a).

They follow language modeling strategies, but aside from text, they also incorporate layout information. Such approaches parse text and layout information by applying OCR on document images. Then, they train transformer-like models. StructuralLM (Li et al., 2021a) adds text and layout embeddings and trains a transformer model (similar to BERT (Devlin et al., 2018)) on specialized pre-training tasks. Some of the recent works employ multi-modal features including visual, text and layout (Xu et al., 2021; Peng et al., 2022; Lee et al., 2021). These models train a single transformer on concatenations of text and visual tokens (Xu et al., 2021) or train a separate transformer branch for both text and visual modalities (Peng et al., 2022). The methods that utilize text consider serialized tokens from OCR as an input, so their performance varies with the correctness of the OCR engine. For examples, if we replace the proprietary Microsoft Azure OCR in LayoutLMv2 (Xu et al., 2021) with Tesseract [1], an open source OCR, its performance drops for visual document classification (Larson et al., 2022).

Transformer-based models consider input sequence based on OCR reading order (Xu et al., 2021; Li et al., 2021a), which may not reflect tokens in their actual reading order (Lee et al., 2021, 2022). Therefore, a few recent studies model the document as a graph by suggesting several possible edge types. Zhang et al. (2020) proposed k-Nearest Neighbors graphs, but these may contain connections with isolated tokens. Fully connected graphs employed by (Liu et al., 2019; Yu et al., 2021) do not leverage the sparsity of the document, hence their approach is similar to transformers. On the other hand, (Cheng et al., 2020) relied on a proprietary OCR technology to identify "text fields", then utilized a 360-degree line-of-sight (LoS) graph. We initially used LoS graphs but that did not show very good performance. FormNet (Lee et al., 2022) models a document as a graph using a $\beta$-skeleton graph (Kirkpatrick and Radke, 1985) and tries to minimize the serialization error by learning localized Super-Token embeddings using graph convolutions before a transformer. However, they used ETC Transformer (Ainslie et al., 2020) for schema learning from GCN-encoded structure-aware Super-Tokens.

Our approach differs from prior graph-based work in two important ways: graph generation and learning embeddings. Our unique document-level sparse graph incorporates **both** spatial layout and OCR reading order, leveraging the document's sparsity and making our model less sensitive to common mistakes in OCR reading order. Moreover, we solely use a GNN to learn embeddings. Thus, we do not require a transformer component, making our approach more memory-efficient than models that incorporate a transformer (Lee et al., 2022; Wei et al., 2020; Yu et al., 2021). Our approach also uses more expressive edge embeddings than that of Liu et al. (2019).

---

[1] https://github.com/tesseract-ocr/tesseract

**Feature fusion** Initial research simply added together the text and layout embedding (Xu et al., 2021; Hong et al., 2022), incorporated position bias in attention mechanism (Garncarek et al., 2021; Powalski et al., 2021), designed cross-modality attention layers (Wang et al., 2022a; Peng et al., 2022; Li et al., 2021b), and explored $1D$ position and $2D$ layout aware attention weights using a disentangled matrix (Peng et al., 2022). LiLT (Wang et al., 2022a) adds attention weights from layout and text embeddings and updates both types of embeddings through two separate transformers. However, adding attention weights does not fully leverage the cross-domain features. SelfDoc (Li et al., 2021b) took the Value (V) of one modality as Key (K) for the other modality while computing cross-attention in transformer layers to learn dependency between language and vision features. Finally, it added features of both text and visual modalities.

## 3 GVdoc Document Graph

We now describe our approach for representing document using both textual and layout features. We represent a document $D$ as a graph where each token is a node and edges reflect the spatial relationship between them.

**Nodes** We define vertices for all tokens as $V = \{v_1, v_2, ..., v_N\}$ where features of $v_i$ are a fusion of the text and layout embeddings defined later in Equation (5). In addition, we define a virtual super node that summarizes the graph, similar to the $CLS$ token in BERT.

**Edges** Token sequence can be important in understanding text, but this information provided by OCR is often noisy. We therefore generate edges in the document graph reflecting two types of relationships between vertices: (a) "ball-of-sight" using $\beta$-skeleton graph (Kirkpatrick and Radke, 1985) and (b) paragraph-based neighborhood.

A $\beta$**-skeleton graph** (Kirkpatrick and Radke, 1985) defines an edge between two bounding boxes if both intersect a circle that does not intersect any other bounding box; the resulting "ball-of-sight" graph is sparser than one using *line*-of-sight edges (Wang et al., 2022b). Lee et al. (2021, 2022) found this useful for message passing in GNNs.

The **paragraph-based neighborhood** connects tokens within the same paragraph and connects paragraphs based on OCR's reading order predictions. While we could fully connect all tokens in the same paragraph, we aim to reduce computation by increasing sparsity; therefore, we add edges for each token with the $k$ nearest neighbors within the same paragraph. Then, for each pair of paragraphs that are adjacent in the OCR's reading order, we define an edge between the last token of the prior paragraph and the first token of the following paragraph. Finally, we define a super-node and connect it with the first and last token of each paragraph, considering them as representative tokens of the paragraph.

To construct the final graph, we take the union of the edges from the $\beta$-skeleton graph and the paragraph-based neighborhood as shown in Figure 1. Thus, we generate a graph that is sparse but also has enough connections for learning node embeddings through message passing in the GNN (as evident in Table 7). For the edge between connected vertices $v_i$ and $v_j$, we define edge features by concatenating (a) distance between all four corners and centers of token bounding boxes of $v_i$ and $v_j$, (b) absolute distance on horizontal and vertical axes, and (c) ratio of height and width.

## 4 GVdoc Model

Our GVdoc model, shown in Figure 2, consists of input embeddings, feature fusion, and task-specific prediction modules. We learn node embeddings in an end-to-end fashion through various unsupervised pre-training tasks. Then, we fine-tune the model for downstream tasks.

### 4.1 Input embedding

**Text embedding:** Our text embedding module is similar to BERT's (Devlin et al., 2018). To get embeddings of text (T), we add token embeddings, token type embeddings, and position embeddings,

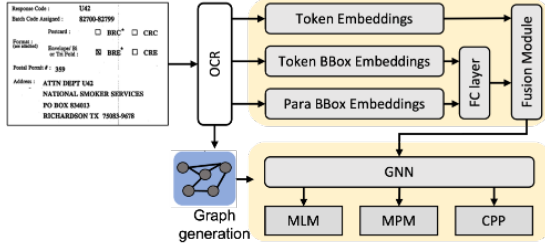

Figure 2: GVdoc overview: OCR returns text tokens, their bounding boxes, and paragraph-level bounding boxes, which are then fed into respective embedding layers. Token and paragraph bounding box embeddings are merged by a fully connected layer and fused with token embeddings through a fusion module. The model is pre-trained on Masked Language Modeling (MLM), Masked Position Modeling (MPM), and Cell Position Prediction (CPP) tasks. Finally, the pre-trained model is fine-tuned for the classification task.

given as

$$e_t = e_{token}(T) + e_{type}(T) + e_{1p}(T) \quad (1)$$

where, $e_{token}, e_{type}, e_{1p}$ are token, token type and position embedding layers, respectively, and $e_t \in R^d$ are text embeddings.

**Layout embedding:** OCR provides text tokens (T), their bounding boxes $T_{box}$, and paragraph-level bounding boxes $P_{box}$. A bounding box contains coordinates of top left corner and bottom right corner, given as $[(x_1, y_1), (x_2, y_2)]$, of a box that covers the token or paragraph. Most document AI models employ token-level bounding boxes for layout embedding that allows the models to localize the text in the layout. StructuralLM (Li et al., 2021a) divides the images into fixed-size grids and uses cell bounding boxes instead of token bounding boxes. They show that the model can encode better contextual information using cell bounding boxes. However, dividing the image into cells might put irrelevant tokens in the same cell or might put a token in two cells. To improve reading order in layout-rich documents, some of the recent approaches (Peng et al., 2022) first detect different text components in the document image and then serialize the tokens from OCR per text component. Motivated by (Peng et al., 2022), we employ text component (paragraph) level layout information for learning layout embeddings. We concatenate the embeddings of paragraph level bounding boxes and token level bounding boxes. Then, we use one fully connected layer to map back to the hidden dimension,

given as:

$$e_l = fc(e_{tl}(T_{box}) \;||\; e_{pl}(P_{box}), \theta) \quad (2)$$

where $||$ denotes concatenation, $e_{tl}$ is a layout embedding layer that encodes token bounding boxes in dimension $R^d$, $e_{pl}$ is a layout embedding layer that encodes paragraph bounding boxes in dimension $R^d$. Finally, both layout embeddings are concatenated to yield a $R^{2d}$ embedding which is mapped into $R^d$ through a fully connected layer. Thus, our layout embeddings $e_l$ contain the coarse and fine-grained location of the tokens based on the document layout.

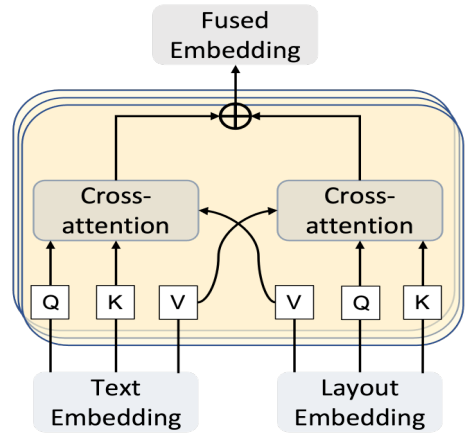

Figure 3: Feature fusion module: Computes cross attention between text embeddings and layout embeddings.

## 4.2 Feature Fusion Module

Our cross-attention module is similar to the cross-attention layer in (Li et al., 2021b), except that we explicitly compute the value representation (V) for both modalities (text and layout) by linear mappings, as shown in Figure 3. Thus, our cross-attention module tries to find the most relevant layout embeddings based on text attention weights and vice versa. Formally we define our cross-attention module in Equation (5).

$$\alpha_t^{ij} = (e_t^i W^{hQ})(e_t^j W^{hK})/\sqrt{d_k} \quad (3)$$
$$\alpha_l^{ij} = (e_l^i W^{hQ})(e_l^j W^{hK})/\sqrt{d_k} \quad (4)$$
$$v_i^h = \sum_{j \in N_i} \alpha_t^{ij}(e_l^i W^{hV}) + \alpha_l^{ij}(e_t^i W^{hV}) \quad (5)$$

where the superscript $h$ represents an attention head, $d_k = d/H$ is the projection dimension (with $H$ being the number of the attention heads), $e_t^i$ and

$e_l^i$ are text and layout embedding vectors fused into node embeddings $v_i^h \in R^{d_k}$ for head $h$. $W^{hQ}$, $W^{hK}$, and $W^{hV}$ are $R^{d \times d_k}$ learnable weights that linearly transform embeddings into queries (Q), keys (K) and values (V), respectively. Node embeddings from all attention heads are concatenated to yield final node embeddings of dimension $d$.

### 4.3 Graph Learning

The generation of document graph results in node features, adjacency matrix and edge features as discussed in Section 3. We chose Graph Attention Network (GAT) (Veličković et al., 2017) as a message passing network for learning node embeddings. The super-node is used to predict the graph (document) label. Our model is first pre-trained in a similar fashion to most of the transformer-based document AI models. We pre-train the model on the following three tasks.

#### 4.3.1 Masked Language Modeling (MLM)

Mask Language Modeling (MLM) is a widely adopted pre-training task in language modeling, involving the masking of random tokens in a text with the special token $MASK$, which the model then aims to predict. Consistent with previous studies (Xu et al., 2021; Li et al., 2021a; Lee et al., 2022), we adopt a masking strategy in which 15% of the tokens are masked. Subsequently, the model learns to estimate the masked tokens based on the information provided by their neighboring tokens.

#### 4.3.2 Masked Position Modeling (MPM)

Each token in the document has its associated location information, represented by a bounding box, which aids in understanding the document's layout. Inspired by the approach presented in Saha et al. (Saha et al., 2021), we randomly replace 15% of the bounding boxes with a fixed bounding box $[0, 0, 0, 0]$. Subsequently, the model is tasked with predicting the masked token-level bounding boxes through a regression task. It is important to note that we do not mask the bounding boxes at the paragraph level, allowing the model to retain access to coarse-grained layout information. As a result, the model's predictions focus solely on the fine-grained layout details while utilizing the provided coarse-grained layout information.

#### 4.3.3 Cell Position Prediction (CPP)

Motivated by (Li et al., 2021a), we divide the document image into a $K \times K$ grid. A token is assigned a cell number in which the center of its bounding box lies. Then, for each token, the model is trained to predict the specific cell within the grid to which it belongs. This task helps the model to narrow down location of tokens within the layout.

## 5 Experiments

We hypothesize that our GVdoc model will be more robust to changes in the test distribution than other models. We therefore designed experiments to measure how our model performed on two tasks: (a) classifying in-domain but out-of-distribution documents, and (b) distinguishing out-of-domain documents from in-domain documents.

### 5.1 Baseline methods

For baseline comparison, we chose models that cover different architectures including CNNs (VGG-16 (Simonyan and Zisserman, 2015), GoogLeNet (Szegedy et al., 2015)), image transformers (DiT (Li et al., 2022), and models that use language modeling (LayoutLMv2 (Xu et al., 2021), LayoutLMv3 (Huang et al., 2022)). Following (Larson et al., 2022), we compare GVdoc with above mentioned models.

### 5.2 Datasets

We use the RVLCDIP (Harley et al., 2015) dataset as our in-distribution and in-domain data, then use RN and RO (Larson et al., 2022) as our out-of-distribution and out-of-domain datasets, respectively.

**RVLCDIP** (Harley et al., 2015) is a subset of IIT-CDIP (Lewis et al., 2006), consisting of scanned and noisy document images from litigation involving the American tobacco industry. The images are labeled for 16 categories including *forms*, *newspaper*, *scientific publication* and so on. The dataset has $320,000$ training samples, and $40,000$ validation and testing examples, each. We fine-tune all models in this work on RVLCDIP's training set. We will use RT to refer to RVLCDIP's test set.

**RVLCDIP-N (RN)** (Larson et al., 2022) is an out-of-distribution but in-domain set. It contains $1,002$ documents belonging to the 12 categories of RVLCDIP dataset, making it in-domain. However, they not taken from the American tobacco industry or IIT-CDIP, so the samples are from a different distribution.

**RVLCDIP-O (RO)** (Larson et al., 2022) was collected from Google and Bing searches and the public Document Cloud [2] repository. It has 3,415 samples, and those documents do not match with any class in RVLCDIP, i.e., they are both out-of-distribution and out-of-domain.

### 5.3 Metrics

**Robustness to out-of-distribution data.** To test how robust each model is to a change in distribution, we compare the model's accuracy on the RVLCDIP test set (RT) and the OOD but in-domain RN. We report both micro-accuracy, calculated as ratio of true positives to total number of samples, and macro-accuracy, calculated by averaging per-class accuracy. A robust model will maintain micro- and macro- accuracy on RN that is close to what it achieved on RT.

**Identifying out-of-domain data.** To test models' effectiveness at identifying out-of-domain data, we follow Larson et al. (2022) in using metrics that describe the separability of confidence scores for in- and out-of- domain examples. A classifier that is good at identifying out-of-domain data should assign high confidence scores to its predictions for in-domain data and low confidence scores to its predictions for out-of-domain data. If we chose a confidence threshold $t$, we could make a binary classifier that labels all examples with confidence $\geq t$ in-domain and all examples with confidence $< t$ out-of-domain; we could then calculate its accuracy, but that accuracy would depend upon our choice of $t$. False positive rate at 95% true positive rate (FPR95) sets $t$ at a level that gives 95% true positives and then measures how many negative examples (out-of-distribution) are classified as positive (in-distribution). A model with a lower FPR95 value model is better at differentiating in- versus out-of-distribution data.

Area under the ROC curve (AUC), similarly, describes how different the confidences are for the in- and out-of-domain examples, but, as a threshold-free measure, is considered as a better option (Larson et al., 2022). A high AUC score (close to 1.0) means the model assigns a higher confidence score to in-domain data and a lower confidence score to out-of-domain data. An AUC score of 0.5 means the model assigns similar confidence scores to in- and out-of-domain samples.

We calculate FPR95 and AUC using two confidence measures: maximum softmax probability and energy score.

**Maximum Softmax Probability (MSP):** Given a model, we compute logits for an example $x$ as $z = f(x)$ and then apply softmax to compute the confidence score per class. For $i^{th}$ class, the confidence score $c_i$ can be calculated as: $c_i = \frac{e^{z_i}}{\sum_j^C e^{z_j}}$, where $C$ is total number of classes. MSP is the maximum confidence score out of these $C$ scores as: $MSP = max\{c_i\}$.

**Energy Score:** Energy score (Liu et al., 2020) is defined as: $E(z, T) = -T \log \sum_{j=1}^{C} e^{(z_j/T)}$ where $T$ is a temperature parameter. For fairness, following (Larson et al., 2022), we use $T = 1$.

### 5.4 Experimental Setup

Given a document, we use OCR to extract text tokens, their bounding boxes, and paragraph (text entity) level bounding boxes. Proprietary OCR engines such as Microsoft Azure OCR used by LayoutLMv2 (Xu et al., 2021), or CLOVA OCR API [3] used by BROS (Hong et al., 2022) are meticulous, but not all users have access to these tools. Thus, following (Larson et al., 2022), we use Tesseract [4], an open source OCR engine, for parsing words and their locations from document images, and then tokenized them using BERT tokenizer. For a better start of training, we initialize text embedding layers with weights from pre-trained BERT.

GVdoc uses an embedding dimension $d = 768$. That is, the dimension for our token embeddings, token bounding-box embeddings and paragraph bounding-box embeddings is $d = 768$. Token and paragraph bounding-box embeddings are concatenated and mapped to final layout embeddings of dimension $d = 768$. Similarly, text and layout embeddings are fused using feature fusion module to result in node embeddings of dimension $d = 768$. Our feature fusion module contains 4 attention heads. We use input edge features of dimension 21, which are also linearly transformed to $d = 768$. We use Graph Attention Network (GAT) (Veličković et al., 2017) with 4 layers and 4 heads. We normalized edge features and input them to GAT.

In our implementation of the $\beta$-skeleton graph (Kirkpatrick and Radke, 1985), we set $\beta = 1$

---

[2] https://www.documentcloud.org
[3] https://clova.ai/ocr
[4] https://github.com/tesseract-ocr/tesseract

| Model | # param | RT | | RN | | Δ RT-RN |
| --- | --- | --- | --- | --- | --- | --- |
| | | Reported | Achieved | Micro | Macro | |
| VGG-16 | 138M | 91.0 | 90.5* | 66.8 | 69.1 | -23.7 |
| GoogLeNet | 60 M | 88.4 | 87.1* | 60.2 | 61.3 | -26.9 |
| DiT | 87 M | 92.1 | 93.3* | 78.6 | 80.5 | -14.7 |
| LayoutLMv2 | 200 M | 95.3 | 88.7* | 55.6 | 60.0 | -33.1 |
| LayoutLMv3 | 133 M | 95.93 | 93.11 | 82.45 | 83.85 | -10.66 |
| GVdoc | 34 M | - | 87.6 | 89.90 | 89.12 | + 2.3 |

Table 1: Classification accuracy scores on RT (Test data) reported by original papers, achieved by (Larson et al., 2022) (indicated by *) compared to RN. Δ RT-RN is the difference in accuracy between RT and RN.

and consider a maximum of 25 neighbors. For the paragraph-level graph, we connect each node to a maximum of 10 nearest neighbors within the same paragraph or text entity, utilizing OCR reading order as the distance metric. We experimented with different numbers of neighbors per text entity, including 5, 10, 15, and 20, but found that selecting 10 neighbors yielded the best performance in terms of accuracy and computational efficiency. Therefore, for all our experiments, we randomly select between 2 to 10 neighbors for each token during training, while during testing, we fix the number of neighbors to 10. The code for GVdoc is publicly available at https://github.com/mohbattharani/GVdoc.

### 5.5 OOD but in-domain performance on RN

Table 1 compares the number of parameters, accuracy on RT reported by their original papers achieved by (Larson et al., 2022), and accuracy on RN (the OOD but in-domain) dataset. Based on the analysis of different models shown in Table 1, almost all previous works reported more than 90% accuracy on the RT except GoogLeNet. More importantly, when these models were tested on the out-of-distribution, in-domain dataset (RN), all the models substantially dropped in accuracy. The original LayoutLMv2 (Xu et al., 2021) utilized the proprietary Microsoft Azur OCR. As a result, when it was evaluated on text parsed using Tesseract OCR, its accuracy on the test set decreased by almost 7%. Furthermore, it performed poorly on the out-of-distribution (OOD) dataset, experiencing a drop of 33% on RN. Notably, the more recent LayoutLMv3 (Huang et al., 2022) exhibited improved performance compared to LayoutLMv2, but it still experienced a drop of nearly 10% on the OOD dataset. DiT appears to have the highest accuracy than the rest on the RT, yet failed to generalize. The drop in accuracy on RN by these models imply that these models might be over-fitting on in-distribution data.

Compared to the top-performing models on the test set, our GVdoc model demonstrates robust performance on RN, indicating its ability to generalize well to out-of-distribution data. Table 2 showcases the per-class accuracy on RN, where GVdoc consistently achieves higher accuracy and accurately categorizes the majority of examples. Notably, our model exhibits high consistency, outperforming or matching the leading results across all classes. In contrast, the other models shows inconsistency, with accuracy dropping below 50% on at least one class. Specifically, for the "Specification" class, our model outperforms all models except LayoutLMv3 (Huang et al., 2022). Moreover, our model achieves nearly 20% higher accuracy than DiT, despite DiT having almost twice the number of parameters as GVdoc. This highlights the effectiveness and efficiency of our model in achieving superior performance.

### 5.6 OOD and out-of-domain results on RO

Here, we compare AUC scores on RT versus RO (T-O), and RN versus RO (N-O) using three metrics: (a) AUC using Maximum Softmax Probability (MSP), (2) AUC using Energy function, and (3) FPR95. These metrics investigate the ability of a model to differentiate between in- and out-distribution data.

**RN vs RO (N-O):** Table 3 compares AUC scores on the out-of-distribution dataset RN versus RO using MSP and energy metrics. The models are trained on the RVLCDIP training set and tested on out-of-distribution datasets − RN and RO. Then their maximum soft-max probability (MSP) and energy function based AUC scores are compared. Ideally, N-O should be more challenging as it compares in-distribution and out-of-distribution datasets (Larson et al., 2022). Among previous approaches, DiT (Li et al., 2022) has the highest test accuracy, and its micro and macro AUC scores using MSP are higher than those of VGG-16, GoogLeNet, and LayoutLMv2. However, our GVdoc model outperforms DiT by 24 points on micro AUC and almost 17 points on macro AUC with MSP. Furthermore, although LayoutLMv3 (Huang et al., 2022) exhibits a test accuracy similar to that of DiT, our model surpasses it. Specifically, GVdoc outperforms LayoutLMv3 by almost 13 points on micro AUC and 9 points on macro AUC with MSP.

Micro- and macro-AUC scores using the En-ergy function do not follow the trend. GoogLeNet achieved the lowest test accuracy and has the lowest Energy AUC scores. Although VGG-16 has higher test accuracy than LayoutLMv2, it is almost 2 points lower on the Micro AUC energy score. Nevertheless, VGG-16 is almost 2 points better on the Macro AUC energy score. DiT and LayoutLMv3 have similar micro and macro scores. GVdoc achieves the highest micro- and macro-AUC scores using energy suggesting that it can effectively differentiate between the in-distribution and out-distribution datasets.

| Model | Micro | Macro | Budget | Email | Form | Handwritten | Invoice | Letter | memo | News Article | Questionnaire | resume | Scientific Pub | Specification |
|---|---|---|---|---|---|---|---|---|---|---|---|---|---|---|
| VGG-16 | 66.8 | 69.1 | 79.3 | 84.8 | 74.3 | 40.3 | 73.7 | 90.1 | 55.3 | 68.6 | 71.8 | 69.6 | 97.4 | 23.0 |
| GoogLeNet | 60.2 | 61.05 | 77.59 | 81.81 | 70.0 | 44.88 | 43.86 | 81.56 | 55.32 | 61.63 | 51.28 | 60.87 | 92.31 | 11.47 |
| DiT | 78.6 | 80.5 | 86.2 | **97.0** | 91.4 | 62.4 | 86 | **95.4** | 72.3 | **84.9** | 82.1 | 73.4 | 92.3 | 41.0 |
| LayoutLMv2 | 55.6 | 60 | 89.7 | 84.8 | 52.9 | 26.1 | 33.3 | 83.6 | 51.1 | 51.2 | 76.9 | 56.5 | 92.3 | 16.4 |
| LayoutLMv3 | 82.4 | 83.8 | 91.2 | 90.62 | **91.9** | 25.7 | 92.3 | 92.5 | 76.2 | 77.8 | **97.8** | 95.3 | **97.9** | **76.8** |
| GVdoc | **89.9** | **89.1** | **98.3** | 82.8 | 89.9 | **85.1** | **96.7** | 95.1 | **87.9** | 81.9 | 97.4 | **97.3** | 97.4 | 61.7 |

Table 2: The per-class accuracy scores on RN (OOD but in-domain dataset) for each document classification model demonstrate the superior performance of GVdoc across various classes. Our model consistently achieves higher accuracy, outperforming or matching the best model on 10 classes and ranking as the second-best on 3 classes.

| Model | MSP | | Energy | |
|---|---|---|---|---|
| | Micro | Macro | Micro | Macro |
| VGG-16 | 0.649 | 0.706 | 0.648 | 0.707 |
| GoogLeNet | 0.592 | 0.679 | 0.587 | 0.689 |
| DiT | 0.728 | 0.780 | 0.753 | 0.792 |
| LayoutLMv2 | 0.620 | 0.717 | 0.643 | 0.716 |
| LayoutLMv3 | 0.755 | 0.807 | 0.755 | 0.807 |
| GVdoc | **0.865** | **0.888** | **0.997** | **0.999** |

Table 3: AUC scores (higher better): RN versus RO.

| Model | MSP | | Energy | |
|---|---|---|---|---|
| | Micro | Macro | Micro | Macro |
| VGG-16 | 0.881 | 0.895 | 0.922 | 0.930 |
| GoogLeNet | 0.838 | 0.859 | 0.847 | 0.869 |
| DiT | 0.893 | 0.902 | 0.888 | 0.902 |
| LayoutLMv2 | 0.842 | 0.875 | 0.849 | 0.891 |
| LayoutLMv3 | 0.817 | 0.889 | 0.817 | 0.889 |
| GVdoc | **0.898** | **0.907** | **0.955** | **0.951** |

Table 5: AUC scores (higher better): RT versus RO.

| Model | MSP | | Energy | |
|---|---|---|---|---|
| | Micro | Macro | Micro | Macro |
| VGG-16 | 0.916 | 0.858 | 0.912 | 0.845 |
| GoogLeNet | 0.947 | 0.869 | 0.943 | 0.845 |
| DiT | 0.847 | 0.704 | 0.843 | 0.685 |
| LayoutLMv2 | 0.932 | 0.848 | 0.939 | 0.847 |
| LayoutLMv3 | 0.839 | 0.618 | 0.834 | 0.611 |
| GVdoc | **0.650** | **0.516** | **0.002** | **0.003** |

Table 4: FPR95 scores (lower better): RN versus RO.

| Model | MSP | | Energy | |
|---|---|---|---|---|
| | Micro | Macro | Micro | Macro |
| VGG-16 | 0.649 | 0.533 | 0.465 | 0.391 |
| GoogLeNet | 0.748 | 0.620 | 0.665 | 0.560 |
| DiT | 0.587 | **0.463** | 0.499 | 0.417 |
| LayoutLMv2 | 0.717 | 0.592 | 0.753 | 0.574 |
| LayoutLMv3 | **0.578** | 0.531 | 0.576 | 0.528 |
| GVdoc | 0.593 | 0.488 | **0.250** | **0.233** |

Table 6: FPR95 scores (lower better): RT versus RO.

Table 4 compares FPR95 scores where a model with lower score is considered better. Micro FPR95 with MSP is in low 0.90's for all the models except LayoutLMv3, DiT and ours. Unlike rest of the models, energy-based FPR95 scores for our model are almost perfect i.e., close to zero. This is evident from the distribution of energy scores in Figure 8 (see Appendix). Overall, GVdoc has lower FPR95 scores compared to the other models. Furthermore, the ROC curves in Figure 5 (see Appendix) confirm that our model can effectively differentiate negative (out-of-distribution) from positive (in-distribution) data. More details are discussed in Appendix A.3.

**RT vs RO (T-O):** Table 5 analyzes the AUC scores of the RT versus out-domain RO data. All models in the study have MSP-based AUC scores ranging from 0.8 to 0.9. While DiT has the highest test accuracy among baselines, its MSP AUC scores are slightly lower than our model. Additionally, DiT falls behind in terms of energy-based AUC scores. Although LayoutLMv3 outperforms its predecessor, LayoutLMv2, in terms of macro MSP and energy scores, it is still unable to surpass DiT. However, GVdoc consistently outperforms all others in the study.

Table 6 presents the FPR95 scores on RT versus RO. In terms of MSP-based FPR95, there is no fixed trend, yet our GVdoc model achieves the second-best FPR95 score based on Macro MSP. In terms of energy-based FPR95, GVdoc outper-forms the rest. VGG-16 achieves a better Micro FPR95 score, whereas GVdoc is $0.146$ points better than VGG-16 in terms of Macro FPR95. Although VGG-16 has lower test accuracy than DiT, its energy-based AUC and FPR95 scores are better than DiT. Overall, GVdoc consistently performs the best in terms of AUC scores and energy-based FPR95, but it is the second-best in MSP-based Macro FPR95.

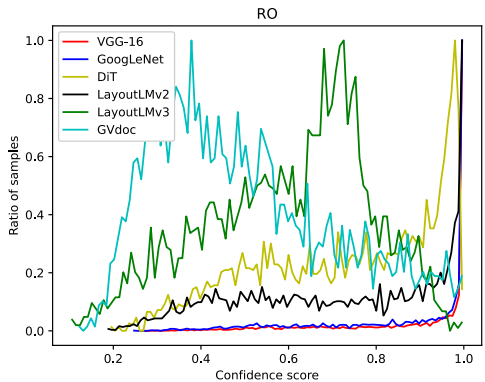

Figure 4: Distribution of confidence scores on RO for different models. GVdoc consistently demonstrates lower confidence scores on out-of-domain data points, indicating its cautious approach towards assigning class labels to unseen classes.

To further investigate this, we plot MSP scores on RO for different models in Figure 4. We can see that our GVdoc model predicts lower confidence scores for out-domain data samples. Figure 6 (see Appendix) demonstrates that the predicted confidence scores for RN and RT are close to $1.0$ for most of the examples. By selecting the proper threshold on confidence scores, we can correctly differentiate between in-domain versus out-domain, and in-distribution versus out-of-distribution data with our model. ROC curves in Figure 5 (see Appendix A.3) show that GVdoc is equivalent or even better than the other models.

### 5.7 Ablation Study

**Effect of graph generation methods** As an ablation study, we compare the effect of different graph generation methods for visual documents. Table 7 demonstrates the importance of the $\beta$ skeleton graph for document classification. Regardless graph generation method, classification accuracy on RT is almost the same. But, using only paragraph-level graphs (based on OCR reading order), the methods struggle to perform well on RN.

| Method | RT Acc | | RN Acc | |
|---|---|---|---|---|
| | Micro | Macro | Micro | Macro |
| $\beta$ skeleton | 87.40 | 87.36 | 87.07 | 86.67 |
| Paragraph-level | 87.15 | 87.11 | 84.90 | 85.53 |
| Both | 87.54 | 87.50 | 89.90 | 89.12 |

Table 7: Comparison of graph generation methods: All the models achieved almost similar classification accuracy on RT. On RN, most of learning is coming from $\beta$ skeleton graph but OCR-based paragraph-level graph helps to improve on out-of-domain RN.

However, our global graph, which combines both $\beta$ skeleton and paragraph-level-graph, achieves the best accuracy on RT and RN.

**Number of the maximum neighbors per token in graph** As discussed in Section 5.4, we discard neighbors from the paragraph-level graph to make it sparse. We constraint maximum degree per node during training. For testing, we select a fixed number of neighbors per token (degree per node). Table 8 demonstrates that reducing the edges during training makes the model robust to the number of neighbors per token. Therefore, our GVdoc model shows the best performance on OOD data.

| Dataset | Number of neighbors | | | |
|---|---|---|---|---|
| | 5 | 10 | 15 | 20 |
| RN | 89.47 | **89.90** | 89.25 | 89.25 |
| RT | 87.30 | 87.60 | 87.26 | 87.32 |

Table 8: The accuracy of the model varies with the numbers of maximum neighbors per token during testing.

## 6 Conclusion

In this paper, we address the limitation of existing visual document classification models by modeling a document as a graph and learning its embeddings using a graph attention network. By defining two types of edges ($\beta$ skeleton and paragraph-based), we leverage the benefit of layout information while minimizing the effects of the errors from OCR reading order. Thus, effectively embracing coarse and fine-grained layout information, GVdoc generalizes better for different layouts. While most visual document classifiers tend to perform well on in-distribution data, they fail or struggle on out-of-distribution data; our model does not drop its performance on OOD data. Through experiments, we demonstrate the generalization of our model on out-of-distribution data.

## 7 Limitations

- We employed Tesseract OCR, an open-source OCR system, which can sometimes make errors in text detection and recognition. However, commercially available OCR engines such as Microsoft Azure OCR are more proficient in detecting text and layout from visual documents. OCR errors can propagate during training and affect the model's performance. For instance, we observed that when Tesseract OCR was used instead of Microsoft Azure OCR, LayoutLMv2 (Xu et al., 2021) experienced a 7% decrease in performance.

- Our model relies on textual and layout features, neglecting the visual component. Various works (Li et al., 2021b; Xu et al., 2021) have already witnessed improvements by utilizing visual features along with textual and layout features. We plan to investigate integration of visual features.

## A Appendix

### A.1 Training Details

We pretrained the model on IITCDIP for one epoch on 64 Tesla V100 GPUs (8 nodes with 8 GPUs per node) with batch size 128 (2 per GPU). We fine tuned the model on RVLCDIP for 100 epochs on 8 GPUs with batch size of 32 (4 per GPU). We used $AdamW$ optimizer with initial learning rate of 0.001 and weight decay of 0.1, for both pre-training and fine-tuning.

### A.2 Ablation Study

**Effect of embedding dimensions:** Table 9 compares the different values for the embedding dimension $d$. The lowest embedding dimension ($d = 128$) does not have enough information for generalization. Comparing the performance on RN vs. RT, we see that using $d = 128$ results in a drop in performance on RN. However, for larger values, starting at $d = 256$, we have see better performance on RN vs. RT. We obtain better scores on RT and OOD RN for $d = 768$. Therefore, $d = 768$ is default embedding dimension for GVdoc.

| $d$ | Micro RT | Macro RT | Micro RN | Macro RN |
|-----|----------|----------|----------|----------|
| 128 | 86.78    | 86.67    | 85.01    | 85.16    |
| 256 | 86.26    | 86.20    | 88.16    | 88.12    |
| 768 | 87.60    | 87.36    | 89.90    | 89.12    |

Table 9: The effect of embedding dimension: Increasing $d$ has a positive impact on generalization.

### A.3 ROC Curve

Figure 5 (left) compares Receiver Operating Characteristic curves (ROC) for in-domain RN versus out-domain RO denoted as (N-O). ROC curve of our GVdoc model is significantly better than the rest of the models. GoogLeNet has AUC score 0.59 and ROC curve close to 0.5 indicates it can not differentiate between in- and out-domain data.

For RT versus RO (T-O), DiT has AUC score of 0.89 whereas our model has 0.9. The ROC curve in Figure 5 (right) demonstrates that GVdoc is close to DiT. Moreover, it has better AUC score and ROC curve than LayoutLMv2 and GoogLeNet for T-O. Overall, GVdoc can effectively differentiate between in-domain and out-of-domain data.

### A.4 Distribution of confidence scores

We plot prediction confidence scores in Figure 6 for RT and RN, respectively. Similar trends suggest that all model have similar confidence score on both datasets. However, Figure 7 demonstrates that our model predicts lower confidence scores on RO suggesting that it is not certain in classifying OOD samples. Whereas, DiT assigns higher confidence to fewer examples, hence incorrectly classifies them into specific class.

### A.5 Distribution of energy scores

When we compare the distribution of energy scores for different models, our GVdoc model has a clear separation between energy scores for RN-RO and RT-RO, as shown in Figure 8. However, from Figure 9, it is hard to differentiate the energy scores of the positive (in-distribution) and negative (out-of-distribution) data samples for the DiT model. The energy scores of VGG-16 for RN and RO in Figure 10 are similar, whereas energy scores for RT-RO are clearly separable.

### A.6 Sample Document Graph

Figure 11 shows an example of a combined graph constructed by merging both the $\beta$ skeleton and OCR-based paragraph-level graph.

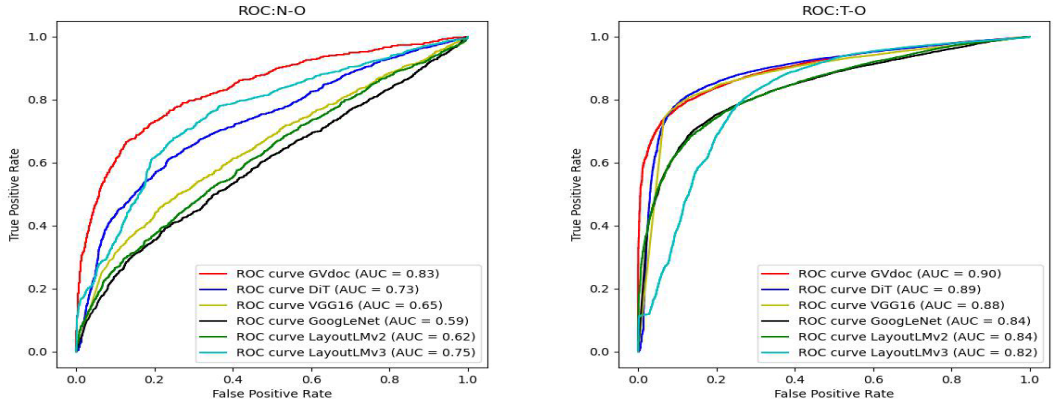

Figure 5: ROC curves for N-O (left) (i.e differentiating RN versus RO examples) and T-O (right) (i.e differentiating RT versus RO examples) suggest that our model is consistently better in differentiating the in-distribution and out-distribution data.

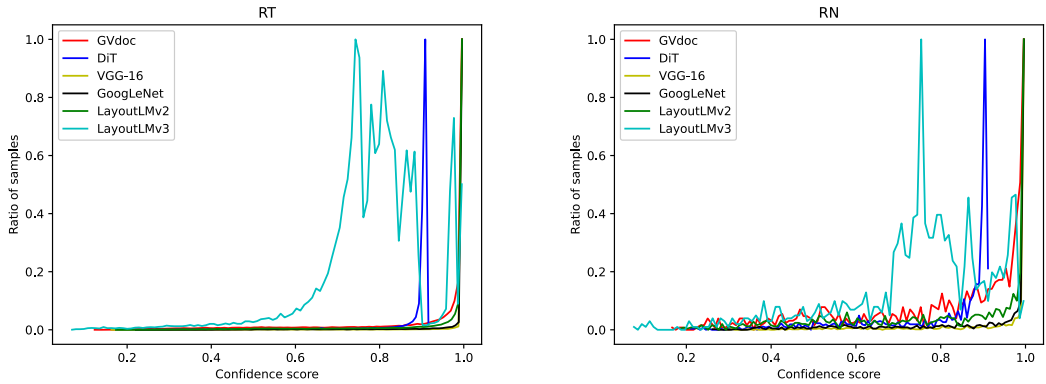

Figure 6: Distribution of confidence scores on RT (left) and RN (right) for different models.

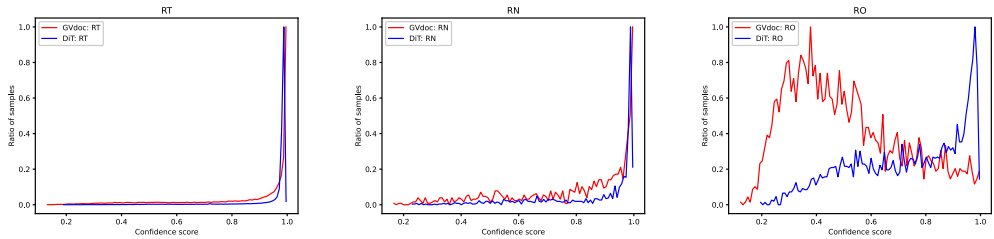

Figure 7: Distribution of confidence scores predicted by out model and DiT on RT, RN, and RO.

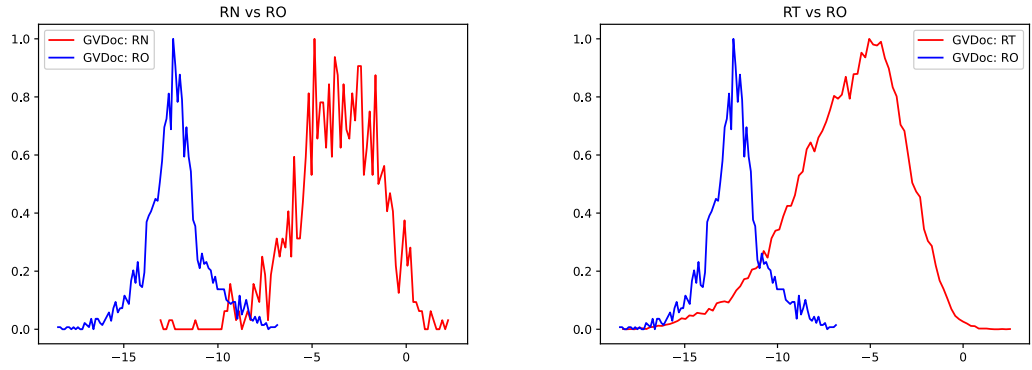

Figure 8: Distribution of energy scores of GVdoc; Left: RN-RO, Right: RT-RO.

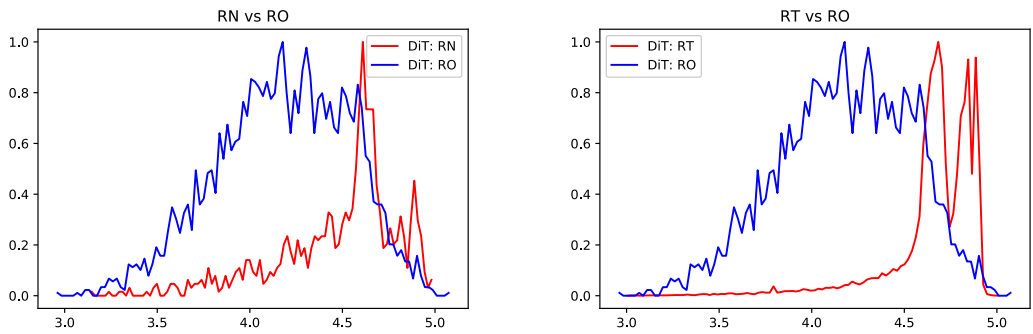

Figure 9: Distribution of energy scores of DiT; Left: RN-RO, Right: RT-RO.

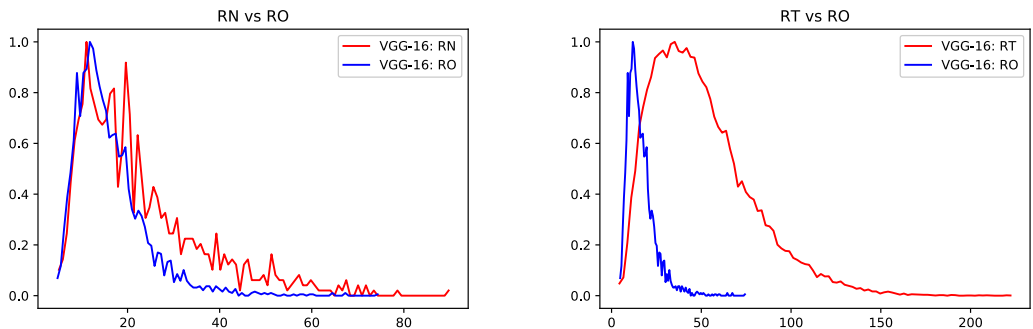

Figure 10: Distribution of energy scores of VGG-16; Left: RN-RO, Right: RT-RO.

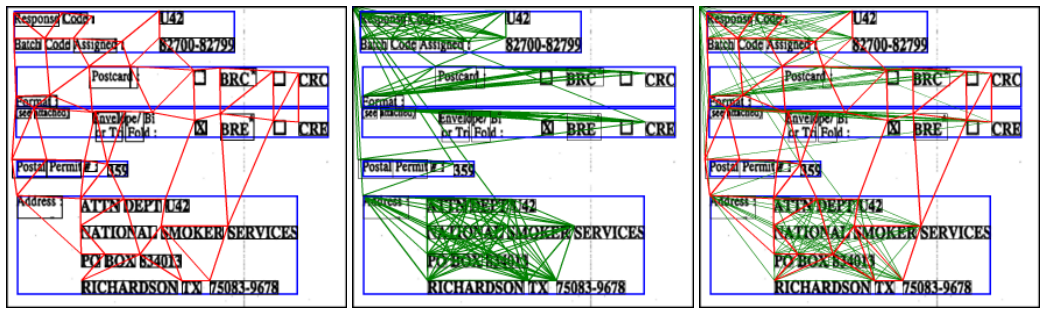

Figure 11: Graph generation: left is $\beta$ graph, middle is OCR-based paragraph-level graph and right is combination of both.